\pdfoutput=1

\documentclass[11pt]{article}
\usepackage{enumitem}
\usepackage{acl}
\usepackage{amsmath}
\usepackage{times}
\usepackage{latexsym}
\usepackage{arydshln}
\usepackage[T1]{fontenc}

\usepackage[utf8]{inputenc}
\usepackage{graphicx}
\usepackage{microtype}

\usepackage{booktabs, multirow} 
\usepackage{soul}
\usepackage[boldmath]{numprint}
\newcommand{\bftab}{\fontseries{b}\selectfont}
\definecolor{myblue}{rgb}{0.9, 0.1, 0.94}
\definecolor{tablegreen}{rgb}{0.82, 0.94, 0.75}
\definecolor{mygreen}{rgb}{0.3, 0.5, 0.78}
\definecolor{myyellow}{rgb}{0.98, 0.94, 0.75}
\newcommand{\metricname}{\textsc{T5Score}}

\title{Multilingual Sequence-to-Sequence Score}
\title{Teach Metrics to Generate and Discriminate}
\title{\textsc{T5Score}: Discriminative Fine-tuning of Generative \\Evaluation Metrics}

 \author{Yiwei Qin$^\clubsuit$ \;\; Weizhe Yuan$^\spadesuit$ \;\; Graham Neubig$^{\clubsuit\heartsuit}$ \;\;  Pengfei Liu$^{\clubsuit\heartsuit}$\thanks{\ \ Corresponding author} \\ 
 \\
  $^\clubsuit$Carnegie Mellon University,  $^\spadesuit$New York University, $^\heartsuit$Inspired Cognition \\ 
  \texttt{\{yiweiq,gneubig,pliu3\}@cs.cmu.edu} \;\;  \texttt{wy885@nyu.edu} \\
  }

\begin{document}
\maketitle
\begin{abstract}
Modern embedding-based metrics for evaluation of generated text generally fall into one of two paradigms: \emph{discriminative} metrics that are trained to directly predict which outputs are of higher quality according to supervised human annotations, and \emph{generative} metrics that are trained to evaluate text based on the probabilities of a generative model.
Both have their advantages; discriminative metrics are able to directly optimize for the problem of distinguishing between good and bad outputs, while generative metrics can be trained using abundant raw text.
In this paper, we present a framework that combines the best of both worlds, using both supervised and unsupervised signals from whatever data we have available.
We operationalize this idea by training \metricname, a metric that uses these training signals with mT5 as backbone\footnote{We use T5 because it supports more languages compared to other options (e.g., BART) and provides different scales of models (e.g., 3B, 11B).}
We perform an extensive empirical comparison with other existing metrics  
on 5 datasets, 19 languages and 280 systems, demonstrating the utility of our method.\footnote{Appendix \ref{sec:appendix-dataset-eval} shows the dataset details.}
Experimental results show that: \metricname achieves the best performance on all datasets against existing top-scoring metrics at the segment level. We release our code and models at 
\url{https://github.com/qinyiwei/T5Score}.

\end{abstract}

\section{Introduction}\label{intro}

Automatically evaluating the quality of generated text plays an essential role in the development of text generation systems~\cite{lin-hovy-2003-automatic,peyrard-2019-studying,mathur-etal-2020-tangled}.
A key element of this evaluation is the design of an automated metric that can \textit{recognize high-quality texts}.
The current most-popular approach to create such high-quality metrics is the \emph{discriminative} paradigm.
These models are generally trained by taking a sentence embedding model and fine-tuning it using human judgments of generated text quality as a learning signal, allowing metrics to directly predict the quality score of a text.
Popular examples include COMET~\cite{rei2020comet} and BLEURT~\cite{sellam2020bleurt}.
However, the effectiveness of this method comes at the cost of expensive manual annotation of human judgements, and thus these models are less broadly applicable than more common lexical metrics such as BLEU~\cite{papineni2002bleu}, ROUGE~\cite{lin2004rouge}.

\begin{figure}[t]
\centering
\includegraphics[width=7.5cm]{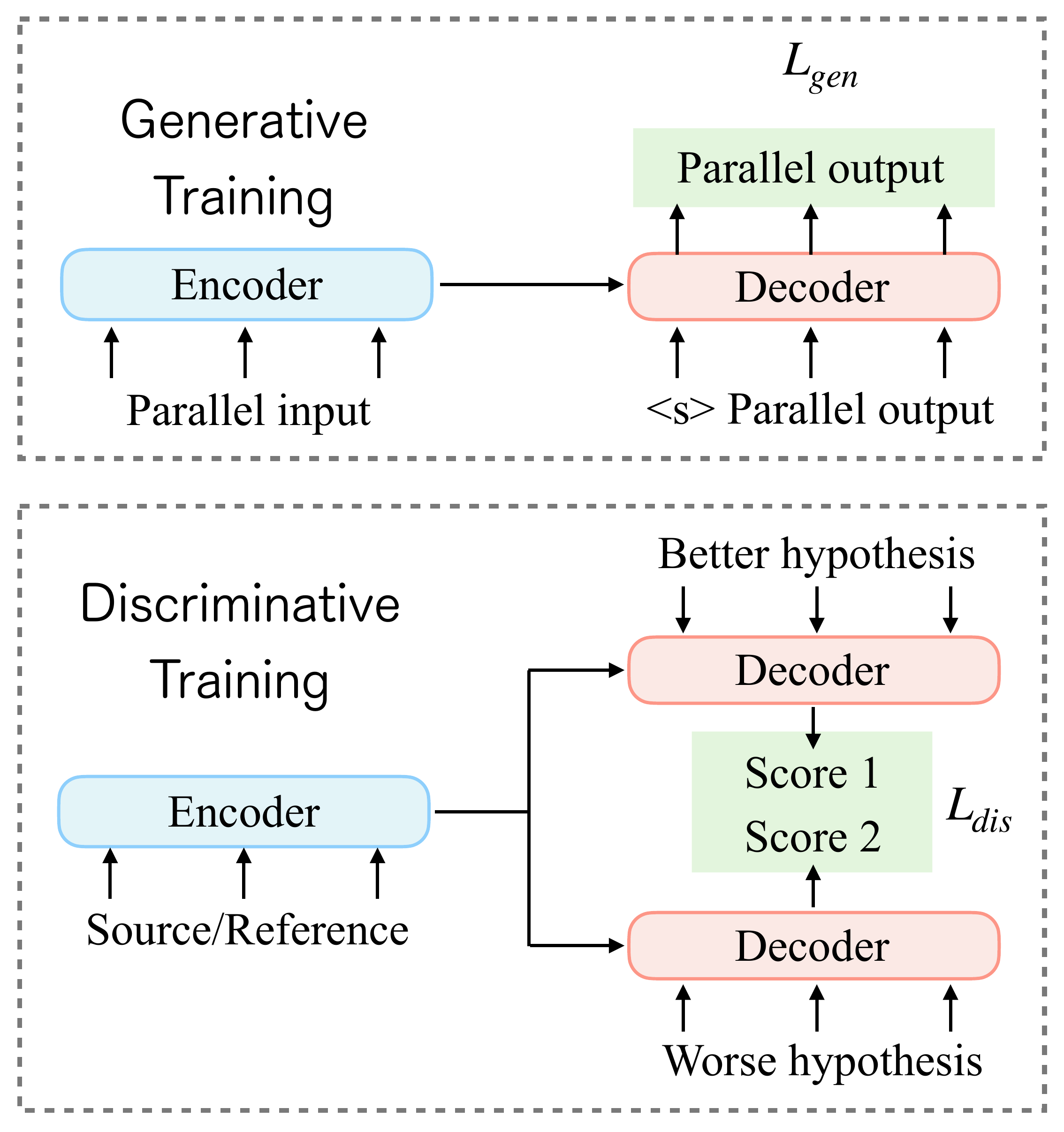}
\caption{Our framework supports generative and discriminative training. The former uses parallel data and maximizes the probability of output data conditioned on the input data. The latter ranks two hypotheses by their manual scores and maximizes the probability of the better hypothesis while minimizing the probability of the worse hypothesis. $L_{gen}$ and $L_{dis}$ denote generative loss and discriminative loss respectively.
} 
\label{model}
\end{figure}

More recently, there has been promising work on \emph{generative} metrics.
These metrics recognize high-quality texts by formulating evaluation as a generation task and using the generative likelihood as an indication of quality, making it possible to train the models without explicit human annotation.
Examples of these metrics include BARTScore \citep{yuan2021bartscore} and PRISM \citep{thompson2020automatic}.
However, because such generative models do not utilize human judgements at training time, these models are inherently at a disadvantage compared to metrics that can utilize supervision.

In this work, we argue that it is crucial to utilize \emph{all} possible supervision symbols that could indicate the quality of the text. 
To this end, we propose a framework for learning evaluation metrics based on the assumption that \textit{generative and discriminative objectives can work in concert to train a better evaluator}, as shown in Fig.~\ref{model}.

We achieve this idea by (1) starting with the pre-trained model mT5~\cite{xue-etal-2021-mt5}, (2) training mT5 in a generative fashion by maximizing the probability of existing parallel data, then (3) fine-tuning mT5 discriminatively by minimizing a contrastive loss function to teach it to ensure that the generative probability of high-quality texts is higher than that of low-quality texts.  At evaluation time,  the probability of generating a text is used as the quality score, because the model has learned to assign high probability to superior texts. Our framework has the flexibility to choose from a supervised training strategy and an unsupervised training strategy depending on if human judgements are available for a given language or task, while keeping the evaluation process the same. 

We evaluate the proposed metric (\metricname) on 5 datasets covering machine translation (MT) and summarization tasks across 19 languages. Regarding reference-based experiments, at the segment level, \metricname trained generatively achieves the best performance on one dataset without human annotated training examples; \metricname trained discriminatively achieves the best performance on 4 datasets with human annotated training examples against top-scoring counterparts. At the system level, \metricname trained discriminatively achieves the best performance in 5 of 6 test settings (2 correlation methods $\times$ 3 datasets).
Empirical results also show the effectiveness of generative training, especially for tasks without human judgments. 
Regarding source-based experiments, we find it better at evaluating top-scoring systems compared to reference-based evaluation, showing the importance of developing source-based evaluation as machines generate higher-quality texts.

\section{Task Formulation}

Text generation evaluation aims to design a function \text{auto\_eval}$(\cdot)$ that takes in a source text $\mathbf{x}$, some reference outputs $\mathbf{y}$ and a system output $\hat{\mathbf{y}}$ and predicts a scalar value that indicates the quality of the system output. The validity of the designed function depends on the degree of correlation between \text{auto\_eval}$(\cdot)$ and human judgements (which can be denoted as \text{manual\_eval}$(\cdot)$). The better the correlation between the two, the more effective we consider our designed function to be.

Specifically, in this work, we call an evaluation function
(1) \textit{source-based} if it takes only $\mathbf{x}$ and $\hat{\mathbf{y}}$ and predicts using \text{auto\_eval}$(\mathbf{x}, \hat{\mathbf{y}})$ (2) and call an evaluation function \textit{reference-based} if it takes only $\mathbf{y}$ and $\hat{\mathbf{y}}$ and predicts using \text{auto\_eval}$(\mathbf{y}, \hat{\mathbf{y}})$, or it takes $\mathbf{x}$, $\mathbf{y}$ and $\hat{\mathbf{y}}$ and predicts using \text{auto\_eval}$(\mathbf{x}, \mathbf{y}, \hat{\mathbf{y}})$.

\section{Metric Design}

In this section, we describe \metricname and explain how to train the metric in both a generative and discriminative fashion.

\subsection{Evaluation as Generation}

Following \citet{yuan2021bartscore}, we formulate text generation evaluation as a text generation problem.

Specifically, the quality of a generated text is measured by calculating the per-token conditional probability of one text $\mathbf{a}$ given another text $\mathbf{b}$, which we also abbreviate as ``$\mathbf{b} \rightarrow \mathbf{a}$'':
\begin{align}
    \label{eq:uniscore}
    \textsc{\metricname} = \frac{1}{|\mathbf{a}|} \log p(\mathbf{a} | \mathbf{b}; \theta) 
\end{align}
$\theta$ are the parameters of the sequence-to-sequence model used to calculate these probabilities.
Depending on which strings we use for $\mathbf{a}$ and $\mathbf{b}$ we can evaluate the text from different perspectives. We adopt the definition of \textit{Precision}, \textit{Recall} and \textit{F} score based on different generation directions \citep{yuan2021bartscore}:

\noindent\textbf{\textit{Precision}: ($\mathbf{x}$ or $\mathbf{y}$ $\rightarrow$ $\hat{\mathbf{y}}$):} 
Calculate probability from reference (or source) text to generated hypothesis  $p(\hat{\mathbf{y}}|\mathbf{y}; \theta)$ (or $p(\hat{\mathbf{y}}|\mathbf{x}; \theta)$).


\noindent\textbf{\textit{Recall} ($\hat{\mathbf{y}}$ $\rightarrow$ $\mathbf{x}$ or $\mathbf{y}$):} 
Calculate probability from generated hypothesis to reference (or source) text $p(\mathbf{y}|\hat{\mathbf{y}}; \theta)$ (or $p(\mathbf{x}|\hat{\mathbf{y}}; \theta)$).

\noindent\textbf{\textit{F} score ($\mathbf{x}$ or $\mathbf{y}$ $\leftrightarrow$ $\hat{\mathbf{y}}$):} The arithmetic average of Precision and Recall to consider both directions. 

According to preliminary experiments, \textit{F} score correlated better with human evaluation scores on the DA20 dataset (\S\ref{sec:eval-datasets})
than \textit{Precision} and \textit{Recall}, so we adopt \textit{F} score for default. In order to support multilingual evaluation, we choose mT5 \citep{xue-etal-2021-mt5} as our pre-trained sequence-to-sequence model.

\subsection{Generative Training for \metricname}
Generative training aims to teach the model to generate target text from the input text with a standard negative log likelihood loss:
 \begin{equation}
     \mathcal{L}_{\text{gen}} = - \frac{1}{m} \sum_{t=1}^m \log p(\mathbf{y}_t | \mathbf{y}_{<t}, \mathbf{x}; \theta).
 \end{equation}

We use the MT dataset ParaCotta \citep{aji2022paracotta} and paraphrasing dataset MT-prism \citep{thompson-post-2020-automatic} as parallel corpora\footnote{Appendix \ref{sec:appendix-dataset-parallel} shows the corpus details.} to train our models generatively.

\subsection{Discriminative Training for \metricname}

We also design discriminative training methods where human judgments for generation quality are available. Suppose we have an annotated training dataset $\mathcal{D} = \{\mathbf{x}_i,\mathbf{y}_i,\hat{\mathbf{y}}_i,m_i|i=1,...,N\}$, where $\mathbf{x}_i$, $\mathbf{y}_i$, $\hat{\mathbf{y}}_i$, and $m_i$ denote the $i$-th example of the source text, the reference text, the hypothesis text, and the manual score, respectively ($\hat{\mathbf{y}}_i$ and $m_i$ can be multiple hypotheses with their corresponding quality scores). We first generate a relative rank dataset $\mathcal{D}_\text{{RR}} = \{\mathbf{x}_i,\mathbf{y}_i,\hat{\mathbf{y}}^+_i,\hat{\mathbf{y}}^-_i,m^+_i,m^-_i|i=1,...,N\}$ by finding a pair of hypotheses $\hat{\mathbf{y}}^+_i$ with higher manual score $m^+_i$ and $\hat{\mathbf{y}}^-_i$ with lower manual score $m^-_i$ for the same source text $\mathbf{x}_i$ and reference text $\mathbf{y}_i$. Then, to encourage the model to assign higher probabilities to the better hypothesis $\hat{\mathbf{y}}^+$. We adopt a contrastive loss function, following previous work \citep{liu2022brio,hopkins2011tuning}: 
\begin{equation}\label{equ:L_sup}
    \mathcal{L}_{\text{dis}} = \text{max}(0, f(\hat{\mathbf{y}}^-) - f(\hat{\mathbf{y}}^+) + \alpha(m^+ - m^-))
\end{equation}
where $\alpha$ is the weight of the margin term. $f$ is defined as $f(\hat{\mathbf{y}}) = \frac{1}{m} \sum_{t=1}^{m}\text{log }p(\hat{\mathbf{y}}_t|\hat{\mathbf{y}}_{<t},\mathbf{y},\theta)$ for reference-based methods, and $f(\hat{\mathbf{y}}) = \frac{1}{m} \sum_{t=1}^{m}\text{log }p(\hat{\mathbf{y}}_t|\hat{\mathbf{y}}_{<t},\mathbf{x},\theta)$ for source-based methods, where $m$ is the number of tokens in $\hat{\mathbf{y}}$.

Because we adopt F score for evaluation by default, our training process also considers two generation directions: from $\mathbf{x}$ or $\mathbf{y}$ to $\hat{\mathbf{y}}$ and from $\hat{\mathbf{y}}$ to $\mathbf{x}$ or $\mathbf{y}$. We augment the training samples by repeating the corpus $\mathcal{D}_\text{RR}$ and changing $\mathbf{x}$ or $\mathbf{y}$ which is originally the model's input to the output and changing $\hat{\mathbf{y}}$ which is originally the model's output to the input. Thus, half of the time we calculate $f(\hat{\mathbf{y}})= \frac{1}{m} \sum_{t=1}^{m}\text{log } p(\mathbf{y}_t|\mathbf{y}_{<t},\hat{\mathbf{y}},\theta)$ for reference based methods, and $f(\hat{\mathbf{y}}) = \frac{1}{m} \sum_{t=1}^{m}\text{log } p(\mathbf{x}_t|\mathbf{x}_{<t},\hat{\mathbf{y}},\theta)$ for source based methods. 

\section{Experimental Setup}\label{exp}

\subsection{Evaluation Datasets}\label{sec:eval-datasets}
We evaluate on 5 datasets: the Direct Assessment (DA) corpus from WMT20 metrics shared task (\textbf{DA20}; \citet{mathur2020results}); datasets obtained by re-annotating the outputs from WMT20 and WMT21 shared task according to the Multidimensional Quality Metrics (MQM) framework (\textbf{MQM20} \& \textbf{MQM21}; \citet{article}); the dataset of WMT20 shared task on Quality Estimation (\textbf{QE20}; \citet{specia2021findings}); and a multilingual summarization dataset (\textbf{MultiSumm}; \citet{koto2021evaluating}). Details in Appendix \ref{sec:appendix-dataset-eval}.





\subsection{Correlation Measurements}
We consider both system-level and segment-level correlations with human judgments when evaluating automated metrics.
\paragraph{System-level}
System-level evaluation calculates average human scores for each generation system to produce a scalar rating for the system performance. We employ the Pearson correlation (sys-p) and Kendall’s Tau correlation (sys-k) as the evaluation measure for system-level metrics.

\paragraph{Segment-level}
\label{eval_seg_level}
Segment-level correlation measures the correlation over segment-level assessments. 
We keep the same setup as in \citet{mathur2020results} converting Direct Assessment (DA) to DA relative rank (DARR) and adopting a Kendall’s Tau-like (seg-k) formulation as the evaluation measure. We adopt the bootstrapping method (p-value < 0.05) \citep{koehn2004statistical,graham2014randomized} for pair-wise significance tests. 

\subsection{Baseline Metrics}
We consider the following baseline metrics for comparison: \textbf{BLEU} \citep{papineni2002bleu} which is the precision of n-grams of the MT output compared to the reference; \textbf{ROUGE} \citep{lin2004rouge} which measures the lexical overlap between the system and reference; \textbf{COMET} \cite{rei2020comet} which is a discriminative metric that uses XLM-RoBERTa to encode source, hypothesis and reference and can be optimised towards different objectives;\footnote{We use the estimator model wmt20-comet-da trained on WMT DA17 to DA19 unless otherwise stated in Sec.\ref{exp}.} \textbf{BERTScore} \citep{zhang2019bertscore} which computes the cosine similarity between the reference and hypothesis tokens’ embeddings based on BERT \citep{devlin2018bert};\footnote{ We use the model: bert-base-multilingual-cased. } \textbf{BLEURT} \citep{sellam2020bleurt} which is a BERT-based regression model trained on synthetic examples and ratings from WMT;\footnote{We use the model BLEURT-20 trained on human ratings from WMT15 to WMT19.} \textbf{PRISM} \cite{thompson2020automatic} which is a generative metric that scores MT system outputs conditioned on their respective human references;\footnote{We use the model trained on machine translation data for 39 language pairs.} \textbf{BARTScore} \citep{yuan2021bartscore} which is a generative metric that uses BART \citep{lewis2019bart} to evaluate the generated text.\footnote{The original version only supports English, so we generalize it to a multilingual version BARTScore by using mBART \citep{liu2020multilingual} and fine-tune it on ParaCotta.}

\section{Reference-based Evaluation}
We consider two tasks in reference-based evaluation: machine translation (DA20, MQM20 and MQM21) and summarization (MultiSumm).
\subsection{Training Details}\label{exp:ref-based}
We consider four different sizes of base models: mT5-B (580M parameters), mT5-L (1.2B parameters), mT5-XL (3.7B parameters), and mT5-XXL (11B parameters). Both generative and discriminative training are considered, with the former based on ParaCotta corpora and the latter based on WMT DA corpora from 2017 to 2019. Our model implementation is based on Huggingface transformers \citep{wolf-etal-2020-transformers} and we adopt the Adafactor \citep{shazeer2018adafactor} optimizer following \citet{DBLP:journals/jmlr/RaffelSRLNMZLL20}. More details of the hyperparameters, training time, computing resources can be found at Appendix~\ref{sec:appendix-hyperparameters}.




\paragraph{DA20} 
Evaluation of the training models is carried out on the DA20 dataset.

\paragraph{MQM}
For the MQM datasets, we consider various discriminative training data, resulting in the following models.\footnote{More training results could be found at Appendix~\ref{sec:appendix-training-data}.} 
\begin{enumerate}[label=(\alph*)]
    \item \metricname-*$^{20}$ is trained on WMT DA corpus from 2017 to 2019.
    \item \metricname-*$^{21}$ is trained on WMT DA corpus from 2017 to 2020.
    \item \metricname-*$^{21}_{\text{mqm}}$ is (b) further trained for 1 additional epoch on MQM20. 
\end{enumerate}
\paragraph{MultiSumm}
Due to the small size of the MultiSumm dataset (135 examples per language pair), we do not undertake the additional training of a model specific to summarization. Instead, we use models trained on ParaCotta and WMT directly.

\subsection{Results}

\begin{figure}[t]
\centering
\caption{System-level Kendall’s Tau and Pearson correlations for the WMT DA20 corpus. Detailed results can be found at Appendix \ref{sec:appendix-WMT-DA-sys}.} 
\includegraphics[width=7.5cm]{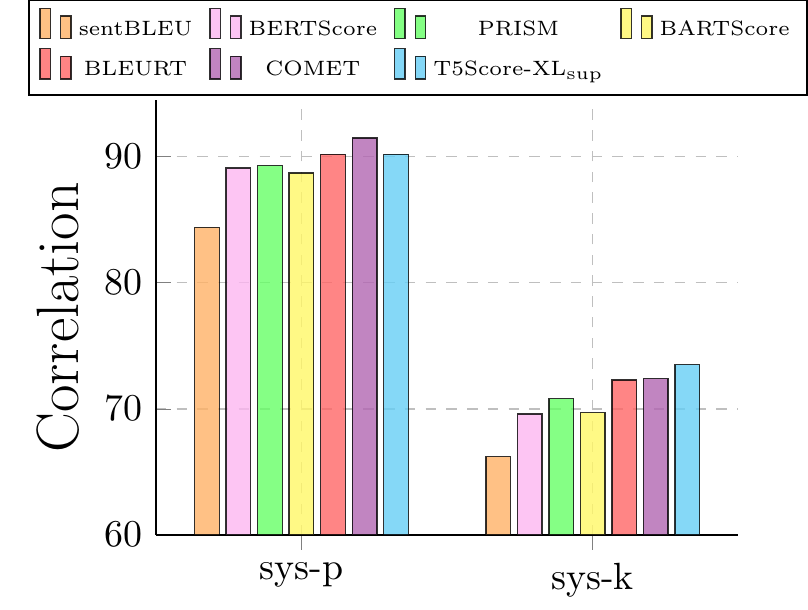}
\label{fig:wmt-da-20-sys}
\end{figure}

\begin{figure}[t]
\centering
\caption{Segment-level Kendall’s Tau correlation on MultiSumm corpus. Details in Appendix \ref{sec:appendix-multi-sum}.} 
\includegraphics[width=7.5cm]{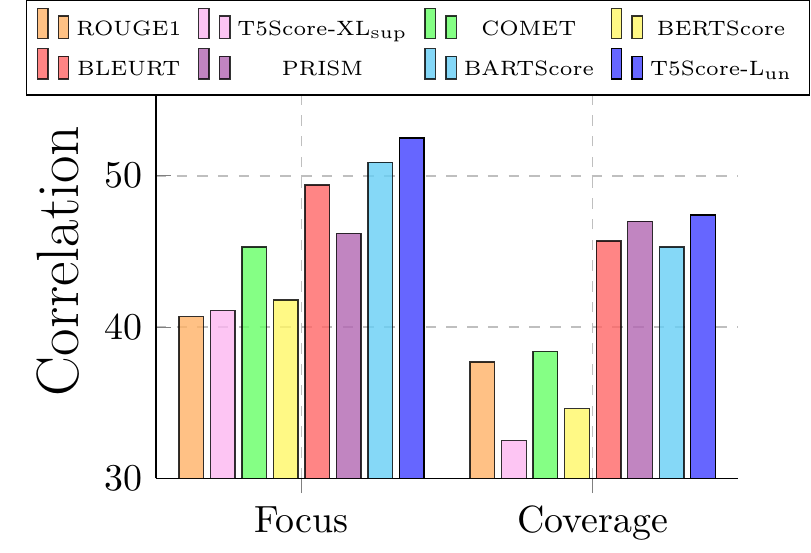}
\label{fig:MultiSumm-seg}
\end{figure}

For DA20, Tab.\ref{tab:wmt-da-20-x-en}/Tab.\ref{tab:wmt-da-20-en-x} shows segment level Kendall’s Tau correlation results of diverse metrics for 10/8 language pairs with English as target/source;\footnote{We find that one MT system NiuTrans.1511 of language pair en-zh generates unnecessary spaces between Chinese characters. Most evaluation metrics cannot handle the spaces well, causing an obvious outlier MT system and influencing correlation greatly, especially Pearson's correlation which is notably sensitive to outliers. To remove the influence, we delete the spaces between Chinese characters of MT system's outputs at evaluation time for every evaluation metric.} Fig.\ref{fig:wmt-da-20-sys} shows system level results on average. For MQM20 and MQM21, Tab.\ref{tab: MQM} shows both segment level and system level results. For MultiSumm, Fig.~\ref{fig:MultiSumm-seg} illustrates the segment Kendall’s Tau correlation.\footnote{For simplicity, 
we report \textit{F} score of our model and other baselines like BERTScore, BARTScore which also differentiate \textit{Precision}, \textit{Recall} and \textit{F} score. 
} \footnote{Because segment scores are reliable only when averaged over sufficient number of judgments \citep{mathur2020results}, we choose Kendall’s Tau-like segment score 
as our correlation metric, different from the original paper \citep{koto2021evaluating}, which uses segment Pearson correlation. We also report Pearson correlation results in Appendix~\ref{sec:appendix-multi-sum} for reference.} 

In all tables, 
\ddag denotes correlations not significantly outperformed by any other metric for the given language pair, while \dag denotes correlations not significantly outperformed by any other unsupervised metric. The highest correlation for each language pair by unsupervised methods is underlined, and the highest correlation overall is bold.

From the above tables and figures,  we observe:

1) \textit{At segment level, our method achieves the best performance on average.} 
Supervised \metricname-XL surpasses all baselines for DA20, MQM20 and MQM21; unsupervised \metricname-L surpasses all baselines for MultiSumm.

2) \textit{At segment level, as language model size increases, the metric performance tends to saturate.} In Tab.\ref{tab:wmt-da-20-en-x}, from \metricname-B to \metricname-L and from \metricname-L to \metricname-XL, the performance of our unsupervised metric improves by 1.4 and 0.6 on average, respectively; while the performance of our supervised metric improves by 1.7 and 0.6, respectively.\footnote{From \metricname-XL to \metricname-XXL, the performance of our unsupervised metrics improves by 0.3. Detailed results can be found at Appendix \ref{sec:appendix-WMT-DA-mt5-xxl}. Due to the limited computational resources and the limited performance improvement of \metricname-XXL, we don't train supervised \metricname-XXL.} However, this is not so clear at system level. A possible reason is that at the system level, there are usually less than 20 systems to be evaluated, much fewer than the number of examples at the segment level, so tiny differences in one MT system can have a large impact on the final results. 

3) \textit{At system level, our method is better at Kendall’s Tau correlation compared to Pearson correlation.} In Fig.\ref{fig:wmt-da-20-sys}, our method achieves the highest Kendall’s Tau correlation compared to other baselines while performs slightly worse than COMET in terms of Pearson correlation. This can be attributed to our training process, which adopts a contrastive loss function, making our models better at predicting the relative rank of examples or systems instead of the absolute score.

4) \textit{For datasets without human annotated training examples, our unsupervised method achieves the best performance.} In Fig.~\ref{fig:MultiSumm-seg}, our supervised methods perform worse than unsupervised methods, and other supervised methods do not work well either. The reason could be that these methods are trained on MT data, and their direct use for the summarization task may impair their performance. These results also indicate that our method has the advantage that the unsupervised version still works well in the absence of human annotated data. 

\begin{table*}[!htp]\centering
\caption{Segment-level Kendall’s Tau correlations on language pairs with English as target for the WMT DA20 corpus. Avg. denotes the average correlation achieved by a metric across all x-en language pairs.
}
\label{tab:wmt-da-20-x-en}
\footnotesize
\begin{tabular}{lrrrrrrrrrrrr}\toprule
&cs-en &de-en &iu-en &ja-en &km-en &pl-en &ps-en &ru-en &ta-en &zh-en &Avg \\\midrule
\multicolumn{12}{l}{\textsc{unsupervised methods}}  \\\midrule
sentBLEU &6.8 &41.1 &18.1 &18.8 &22.6 &-2.5 &9.6 &-0.5 &16.3 &9.3 &13.9 \\
BERTScore &11.7 &45.2 &21.6 &24.3 &27.9 &4.7 &15.9 &6.0 &21.9 &13.4 &19.3 \\
PRISM &\underline{13.5}\ddag &46.5 &25.5 &26.3 &30.4 &6.6 &16.5 &10.0 &23.0 &14.5 &21.3 \\
BARTScore &12.4 &48.5\dag &23.5 &26.6 &\underline{\bftab31.8}\ddag &9.1\dag &16.0 &12.8\dag &23.8 &16.3\dag &22.1 \\\midrule
\metricname-B$_{\text{un}}$ &12.9 &48.4 &24.3 &26.0 &30.4 &8.3 &\underline{\bftab19.4}\ddag &11.9 &\underline{24.0}\dag &15.9 &22.2 \\
\metricname-L$_{\text{un}}$ &13.0\dag &48.7\dag &\underline{26.6}\dag &\underline{27.9}\dag &30.9 &8.5 &17.7 &12.8 &\underline{24.0}\dag &\underline{16.5}\dag &\underline{22.7} \\
\metricname-XL$_{\text{un}}$ &13.0 &\underline{48.8}\dag &26.1\dag &27.7\dag &29.8 &\underline{9.2}\dag &17.7 &\underline{13.2}\dag &23.8\dag &15.9 &22.5 \\\midrule\midrule
\multicolumn{12}{l}{\textsc{supervised methods}} \\\midrule
BLEURT &13.6\ddag &47.6 &27.1 &28.1 &31.2\ddag &4.7 &18.4\ddag &10.3 &25.3 &14.7 &22.1 \\
COMET &12.9 &48.5 &28.1 &27.4 &29.8 &\bftab9.9\ddag &15.8 &\bftab15.6\ddag &24.2 &\bftab17.1\ddag &22.9 \\\midrule
\metricname-B$_{\text{sup}}$ &13.9\ddag &48.5 &\bftab29.2\ddag &28.1 &30.3 &9.6\ddag &17.4 &13.1 &23.9 &15.5 &22.9 \\
\metricname-L$_{\text{sup}}$ &\bftab14.0\ddag &49.3\ddag &28.5\ddag &28.9\ddag &30.1 &8.3 &17.6 &15.3\ddag &\bftab25.9\ddag &16.3 &23.4 \\
\metricname-XL$_{\text{sup}}$ &12.8 &\bftab49.6\ddag &\bftab29.2\ddag &\bftab29.1\ddag &31.5\ddag &9.3 &18.0 &15.2\ddag &25.4 &15.8 &\bftab23.6 \\
\bottomrule
\end{tabular}
\end{table*}

\begin{table*}[!htp]\centering
\caption{Segment-level Kendall’s Tau correlations on language pairs with English as source for the WMT DA20 corpus. Avg. denotes the average correlation achieved by a metric across all en-x language pairs, and Avg-all denotes the average correlation across all language pairs (including en-x and x-en). 
}\label{tab:wmt-da-20-en-x}
\footnotesize
\begin{tabular}{lrrrrrrrrrrr}\toprule
&en-cs &en-de &en-iu &en-ja &en-pl &en-ru &en-ta &en-zh &Avg &Avg-all \\\midrule
\multicolumn{11}{l}{\textsc{unsupervised methods}} \\\midrule
sentBLEU &43.2 &30.2 &19.1 &47.9 &15.3 &5.1 &39.5 &39.7 &30.0 &21.1 \\
BERTScore &51.1 &39.5 &19.5 &53.8 &28.5 &20.5 &60.4 &41.1 &39.3 &28.2 \\
PRISM &\underline{61.2}\dag &\underline{43.7}\dag &19.6 &57.8 &\underline{39.9}\dag &26.8\dag &39.3 &\underline{47.3}\dag &41.9 &30.5 \\
BARTScore &55.9 &42.0 &26.0 &57.9 &32.1 &24.4 &62.8 &46.2 &43.4 &31.6 \\\midrule
\metricname-B$_{\text{un}}$ &52.4 &39.4 &35.8\ddag &55.9 &31.9 &22.4 &62.1 &42.8 &42.8 &31.3 \\
\metricname-L$_{\text{un}}$ &56.3 &41.1 &\underline{\bftab35.9}\ddag &57.4 &35.3 &25.3 &65.9 &45.4 &45.3 &32.7 \\
\metricname-XL$_{\text{un}}$ &58.9 &42.2 &34.0 &\underline{59.8}\dag &38.3 &\underline{27.4}\dag &\underline{66.3}\dag &47.1\dag &\underline{46.7} &\underline{33.3} \\\midrule\midrule
\multicolumn{11}{l}{\textsc{supervised methods}} \\\midrule
bleurt &\bftab69.3\ddag &45.9 &32.6 &60.8 &45.5 &30.1 &65.3 &50.3 &50.0 &34.5 \\
comet &66.8 &\bftab46.8\ddag &32.2 &62.4 &\bftab46.2\ddag &\bftab34.5\ddag &67.1 &\bftab52.3\ddag &51.0 &35.4 \\\midrule
\metricname-B$_{\text{sup}}$ &61.2 &43.1 &32.4 &60.5 &38.9 &28.9 &64.1 &48.5 &47.2 &33.7 \\
\metricname-L$_{\text{sup}}$ &66.6 &45.6 &33.4 &62.8\ddag &43.9 &31.8 &66.7 &52.2\ddag &50.4 &35.4 \\
\metricname-XL$_{\text{sup}}$ &68.0 &\bftab46.8\ddag &34.1 &\bftab63.0\ddag &45.6 &33.8\ddag &\bftab68.1\ddag &51.7\ddag &\bftab51.4 &\bftab36.0 \\
\bottomrule
\end{tabular}
\end{table*}

\begin{table*}[!htp]\centering
\setlength\tabcolsep{3.2pt}
\caption{Segment Kendall’s Tau, system Pearson and system Kendall’s Tau of different metrics on MQM20 and MQM21 dataset. Avg. denotes the average correlation achieved by a metric across two language pairs and two years. Method COMET uses model wmt20-comet-da and wmt21-comet-mqm for MQM20 and MQM21 respectively. Method \metricname-XL$_{\text{sup}}$ uses model \metricname-*$^{20}$ and \metricname-*$^{21}_{\text{mqm}}$ for MQM20 and MQM21 respectively.}\label{tab: MQM}
\footnotesize
\begin{tabular}{lrrrrrrrrrrrrrrrr}\toprule
&\multicolumn{6}{c}{MQM-2020} &\multicolumn{6}{c}{MQM-2021} &\multicolumn{3}{c}{\multirow{2}{*}{avg}} \\\cmidrule(lr){2-7}\cmidrule(lr){8-13}
&\multicolumn{3}{c}{en-de} &\multicolumn{3}{c}{zh-en} &\multicolumn{3}{c}{en-de} &\multicolumn{3}{c}{zh-en} & & & \\\cmidrule(lr){2-4}\cmidrule(lr){5-7}\cmidrule(lr){8-10}\cmidrule(lr){11-13}\cmidrule(lr){14-16}
&sys-p &sys-k &seg-k &sys-p &sys-k &seg-k &sys-p &sys-k &seg-k &sys-p &sys-k &seg-k &sys-p &sys-k &seg-k \\\midrule
\multicolumn{16}{l}{\textsc{unsupervised methods}}  \\\midrule
sentBLEU &82.8 &52.4 &11.3 &43.8 &57.1 &7.6 &88.0 &82.1 &2.8 &35.4 &28.2 &1.5 &62.5 &54.9 &5.8 \\
BERTScore &79.1 &42.9 &20.4 &51.5 &35.7 &15.2 &88.6 &82.1 &11.6 &48.7 &33.3 &5.3 &67.0 &48.5 &13.1 \\
PRISM &\underline{\bftab98.9} &81.0 &\underline{27.8} &\underline{77.8} &\underline{64.3} &\underline{23.3} &80.7 &56.4 &12.7 &49.0 &30.8 &10.0 &\underline{76.6} &58.1 &18.5 \\
BARTScore &91.9 &\underline{\bftab90.5} &25.0 &54.0 &50.0 &20.8 &\underline{86.7} &\underline{\bftab79.5} &\underline{17.9} &43.2 &30.8 &9.5 &69.0 &62.7 &18.3 \\\midrule
\metricname-B$_{\text{un}}$ &94.6 &71.4 &23.4 &52.2 &42.9 &20.2 &82.8 &64.1 &13.3 &49.0 &35.9 &10.4 &69.7 &53.6 &16.8 \\
\metricname-L$_{\text{un}}$ &95.1 &81.0 &25.6 &57.5 &42.9 &21.8 &84.2 &64.1 &15.1 &51.4 &35.9 &10.4 &72.0 &56.0 &18.2 \\
\metricname-XL$_{\text{un}}$ &93.4 &81.0 &27.7 &66.6 &\underline{64.3} &22.7 &86.4 &71.8 &16.2 &\underline{51.7} &\underline{41.0} &\underline{10.5} &74.5 &\underline{64.5} &\underline{19.3} \\\midrule
\multicolumn{16}{l}{\textsc{supervised methods}}  \\\midrule
BLEURT &95.5 &71.4 &30.7 &\bftab91.6 &\bftab78.6 &24.3 &79.7 &64.1 &16.4 &47.3 &41.0 &10.3 &78.5 &63.8 &20.4 \\
COMET &96.5 &71.4 &27.7 &88.9 &71.4 &22.1 &77.1 &66.7 &19.8 &55.9 &33.3 &\bftab14.1 &79.6 &60.7 &20.9 \\\midrule
\metricname-B$_{\text{sup}}$ &98.4 &71.4 &25.5 &65.2 &42.9 &21.8 &\bftab91.7 &\bftab79.5 &18.0 &56.5 &46.2 &11.5 &78.0 &60.0 &19.2 \\
\metricname-L$_{\text{sup}}$ &98.5 &81.0 &29.6 &85.5 &64.3 &25.1 &88.9 &74.4 &\bftab20.7 &58.9 &48.7 &13.6 &83.0 &67.1 &\bftab22.3 \\
\metricname-XL$_{\text{sup}}$ &\bftab98.9 &\bftab81.0 &\bftab31.8 &89.0 &\bftab78.6 &\bftab25.5 &86.0 &69.2 &19.2 &\bftab62.9 &\bftab53.8 &12.5 &\bftab84.2 &\bftab70.7 &\bftab22.3 \\
\bottomrule
\end{tabular}
\end{table*}

\section{Source-Based Evaluation}
We support both reference-based and source-based discriminatively trained methods. In this section, we show the effectiveness of the source-based method. We consider the task of machine translation and inspect the results on three datasets: DA20, MQM21 and QE20.

\subsection{Training details}
Hyperparameters are kept the same as in Sec.~\ref{exp:ref-based}. 
\noindent\textbf{DA20} For DA20, generative training is performed on MT-prism, while discriminative training is performed on the WMT DA corpus from 2017 to 2019.
\noindent\textbf{MQM21} We take the models from DA20 and further train them for 2 additional epochs on MQM20.
\noindent\textbf{QE20} 
We take the models from DA20 and further train them on the QE20 train split. The best checkpoint is picked based on its performance on QE20 development split and the results on the test split are reported.

\subsection{Results}


For MQM21, We compare our supervised model with the COMET$^{\text{mqm}}_{\text{src}}$ baseline \citep{rei2021references}.\footnote{COMET$^{\text{mqm}}_{\text{src}}$ is a source-based metric trained on WMT DA scores from 2017 to 2020 and adapted to MQM by fine-tuning on MQM20. We use the model wmt21-comet-qe-mqm.} Tab.\ref{tab: mqm-src} illustrates both the segment level and system level results. For DA20, we compare our supervised model with COMET$_{\text{src}}$ baseline \citep{rei2021references}.\footnote{COMET$_{\text{src}}$ is trained on WMT DA scores from 2017 to 2019. We use the model wmt20-comet-qe-da.} Results can be found at Appendix \ref{sec:appendix-WMT-DA-src-based}. For QE20, we choose PRISM$_\text{qe}$ \citep{thompson2020automatic} which is a reference free version of PRISM as the unsupervised baseline, and TransQuest \citep{transquest:2020a,transquest:2020b} which is the winner of WMT2020 Shared Task on Quality Estimation \citep{specia2021findings} as the supervised baseline.\footnote{We use the \href{https://github.com/TharinduDR/TransQuest}{Implementation} and the model \href{https://huggingface.co/TransQuest/monotransquest-da-multilingual}{monotransquest-da-multilingual}.} Tab.~\ref{tab: QE} illustrates the segment Pearson correlation\footnote{The correlation metric follows \citet{specia2021findings}.} of different evaluation metrics.

We have the following observations:

1) For MQM21, \metricname surpasses the baseline at both segment and system level on average. For DA20, \metricname is better than the baseline at the segment level on average, and better than the baseline for most language pairs at the system level. For QE20, both supervised \metricname and unsupervised \metricname surpass the corresponding baselines.

2) \textit{Overall, source-based models perform worse than the reference-based models, but their differences are much smaller on MQM21 than DA20.} We conjecture that the reason for this may be related to the human evaluation process where WMT-DA uses a mixture of source-based and reference-based annotations, while MQM uses the source.

\begin{table}[!htp]\centering
\setlength\tabcolsep{2.4pt}
\caption{Segment Kendall’s Tau, system Pearson and system Kendall’s Tau of different metrics on the MQM21 dataset for source-based methods. The highest correlation for each language pair under each correlation method is bold.}\label{tab: mqm-src}
\footnotesize
\begin{tabular}{lrrrrrrr}\toprule
&\multicolumn{3}{c}{en-de} &\multicolumn{3}{c}{zh-en} \\\cmidrule{2-7}
&sys-p &sys-k &seg-k &sys-p &sys-k &seg-k \\\midrule
COMET$^{\text{mqm}}_{\text{src}}$ &68.5 &46.2 &16.2 &50.5 &41.0 &\textbf{10.0} \\
\metricname-B$^{\text{mqm}}_{\text{src}}$ &42.9 &23.1 &13.0 &53.0 &41.0 &6.2 \\
\metricname-L$^{\text{mqm}}_{\text{src}}$ &67.6 &48.7 &16.9 &\textbf{60.8} &\textbf{51.3} &8.6 \\
\metricname-XL$^{\text{mqm}}_{\text{src}}$ &\textbf{77.8} &\textbf{56.4} &\textbf{17.9} &58.7 &48.7 &9.3 \\
\bottomrule
\end{tabular}
\end{table}

\begin{table}[!htp]\centering
\setlength\tabcolsep{1.8pt}
\caption{Segment level Pearson correlation on QE20 corpus for source-based methods. The highest correlation by unsupervised method is underlined, and the highest correlation overall is bold. }\label{tab: QE}
\footnotesize
\begin{tabular}{lrrrrrrrr}\toprule
&si-en &ne-en &et-en &ro-en &en-de &en-zh &Avg \\\midrule
\multicolumn{8}{l}{\textsc{unsupervised methods}}  \\\midrule
PRISM$_\text{qe}$ &2.9 &-13.6 &\underline{69.4} &\underline{82.9} &\underline{46.4} &\underline{30.4} &36.4 \\
\metricname-B$_\text{{u}}$ &53.0 &55.7 &58.1 &77.1 &13.5 &21.3 &46.4 \\
\metricname-L$_\text{{u}}$  &\underline{59.3} &60.4 &64.1 &80.0 &20.4 &25.9 &51.7 \\
\metricname-XL$_\text{{u}}$  &58.8 &\underline{61.7} &64.0 &80.7 &26.7 &28.5 &\underline{53.4} \\\midrule\midrule
\multicolumn{8}{l}{\textsc{supervised methods}}  \\\midrule
TQ &58.9 &\bftab75.5 &76.6 &\bftab88.0 &42.3 &44.1 &64.2 \\
\metricname-B$_\text{{s}}$ &58.1 &69.9 &71.8 &83.4 &44.7 &41.4 &61.6 \\
\metricname-L$_\text{{s}}$ &58.9 &\bftab75.5 &76.8 &86.4 &\bftab49.8 &44.0 &65.2 \\
\metricname-XL$_\text{{s}}$ &\bftab60.2 &74.7 &\bftab77.8 &87.1 &\bftab49.8 &\bftab45.6 &\bftab65.9 \\
\bottomrule
\end{tabular}
\end{table}

\section{Analysis}
We design analyses to better understand the mechanism of \metricname and its strenghs over other metrics, specifically asking three questions: Q1: \textit{Is generative training necessary before discriminative training?} Q2: \textit{What are the strengths and weaknesses of each evaluation metric?} Q3: \textit{When will source-based evaluation outperforms reference-based evaluation?}

\begin{figure}[t]
\centering
\caption{Segment Kendall’s Tau, system Pearson and system Kendall’s Tau correlations of different models with and without generative training on DA20. \metricname-B$_{\text{w}}$ and \metricname-L$_{\text{w}}$ are models with generative training, while \metricname-B$_{\text{w/o}}$ and \metricname-L$_{\text{w/o}}$ are models without generative training.} 
\includegraphics[width=7.5cm]{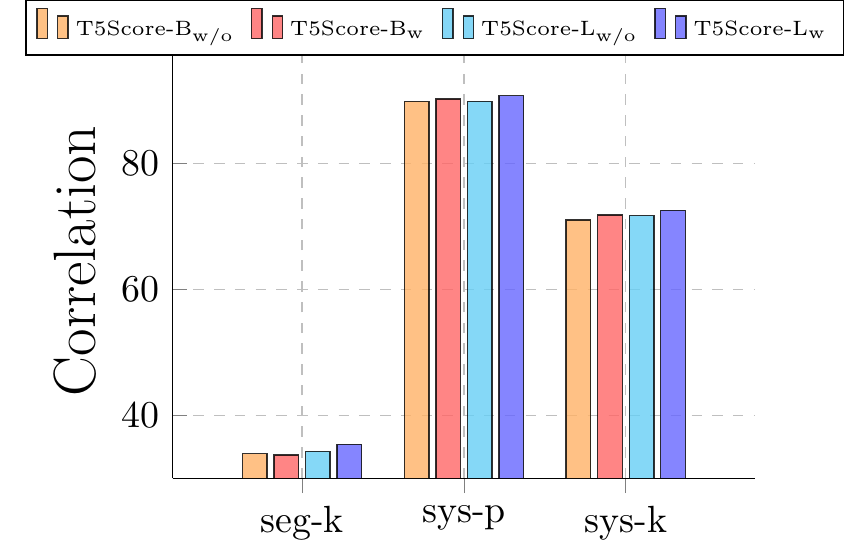}
\label{fig:finetuning}
\end{figure}

\paragraph{Effectiveness of Generative Training}
In our experiments, discriminative training is based on the model trained generatively. To answer Q1, we compare the performance of discriminatively trained models with and without generative training on DA20. The results are shown in Fig.~\ref{fig:finetuning}.\footnote{ Appendix \ref{sec:appendix-WMT-DA-ablation} has detailed results of every language pair.}
We observe that (1) Both \metricname-B and \metricname-L are enhanced by generative training before discriminative training under three correlation metrics (except that \metricname-B using segment level Kendall’s Tau correlation has a little performance drop), which means that generative training is necessary to improve model performance. (2) Larger model \metricname-L benefits more from generative training, compared to \metricname-B, indicating that larger models are better at keeping knowledge from generative training. 

\begin{figure*}[t]
\centering
\caption{RMSE of different metrics in each error category on MQM dataset.} 
\includegraphics[width=15.5cm]{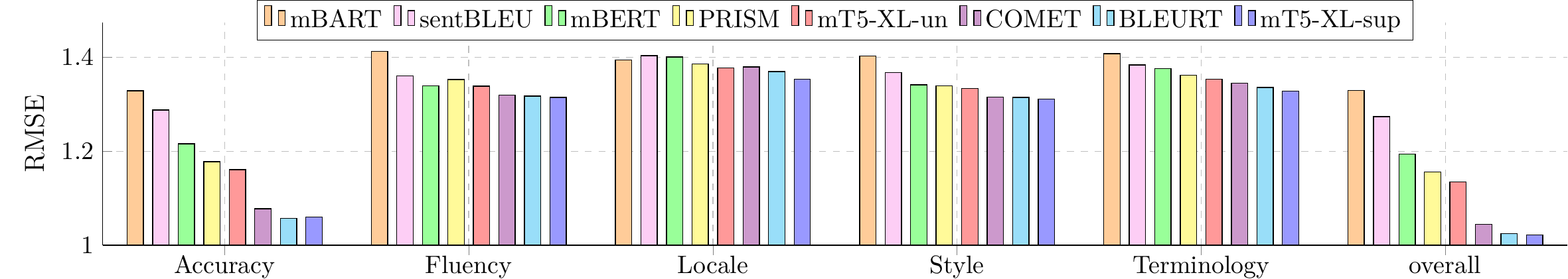}
\label{multi_dim}
\end{figure*}

\paragraph{Multi-Dimension Analysis} 
For Q2, we compare diverse evaluation metrics under different error categories on the MQM datasets. 
To evaluate the performance of the metrics on a given error category, we use the score of each example in that error category as the gold standard to compare with the score given by the automated evaluation metrics. There are six error categories in total, including the five error categories described in Sec.~\ref{sec:eval-datasets} and an overall category that measures all errors.
Fig.~\ref{multi_dim} shows the Root Mean Square Error (RMSE)\footnote{We use RMSE instead of seg-k, because a certain error category only has a few examples, so the number of examples to calculate seg-k for that category will be too small to be accurate. Scores are normalized before calculating RMSE.} of diverse evaluation metrics under different error categories. We observe that: 
(1) Our model \metricname-XL$_{\text{sup}}$ ranks first overall in every error category except accuracy where BLEURT ranks first. 
(2) Supervised metrics (BLEURT, COMET, \metricname-XL$_{\text{sup}})$ perform better than other unsupervised metrics.
(3) The evaluation perspectives that all metrics excel at are the same. All metrics perform best in terms of accuracy, much better than other error categories.

\begin{figure*}[t]
\centering
\caption{Metrics performance of the top-k MT systems. X-axis is number k. Y-axis is correlation results.} 
\includegraphics[width=15cm]{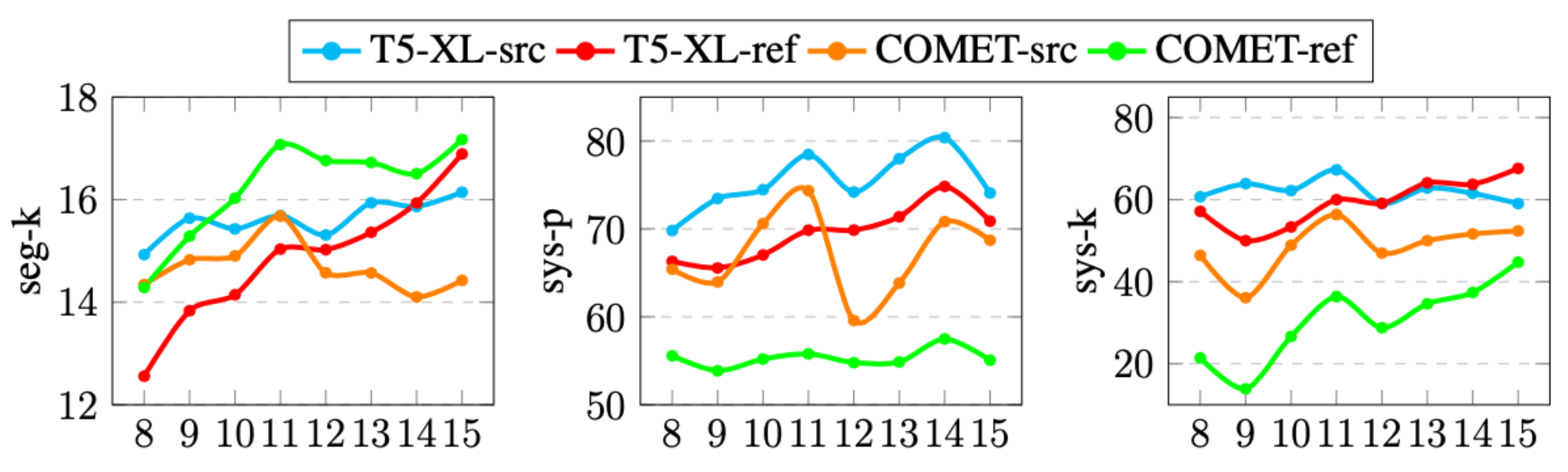}
\label{top-k-sys}
\end{figure*}

\paragraph{Top-k Analysis} 
To answer Q3, we conduct experiments on MQM21 and evaluate on the subset of the data from the top-k performing MT systems.\footnote{To get more high quality translations, we also evaluate human translations in this experiment. All other experiments in this paper don't include human translations.} Results are presented in Fig.~\ref{top-k-sys}. We find that: (1) The advantage of the source-based version of \metricname over the reference-based version increases as we evaluate fewer systems, i.e., only high-quality systems. Although not as pronounced in COMET, which also uses the source for its reference-based approach, it has roughly the same trend. This suggests that the source-based approach is more suitable for evaluating top systems, and that source-based evaluation should be considered as machine systems are improved. (2) \metricname outperforms COMET on all top-k systems under all three correlation measures, except for the reference-based version under seg-k correlation. The better performance of COMET's reference-based version may be attributed to its combination of reference and source, further indicating the importance of source.


\section{Related Work}\label{references}


The increasing performance of generation systems equipped with large pre-trained models puts forward a higher requirement of the evaluation ability of automated metrics.
As such, researchers are exploring different evaluation frameworks by 
teaching metric to learn diverse types of knowledge.

The directest way is to supervise metrics with manually annotated judgments with a trainable model, typical works include BEER \citep{stanojevic2014beer}, BLEURT \citep{sellam2020bleurt}, ROBLEURT \citep{wan2022robleurt}, COMET \citep{rei2020comet} and C-SPEC \citep{takahashi2021multilingual}. 

Despite the superior performance of optimizing the correlation with human judgments, this method is expensive in creating human annotations.
To bypass this challenge, researchers attempt to evaluate generated texts in an unsupervised way by calculating the lexical or semantic similarity between reference and generated texts with surface-based string match (e.g., BLEU \citep{papineni2002bleu}, ROUGE \citep{lin2004rouge} and CHRF \citep{popovic2015chrf}) or unsupervised pre-trained components (e.g., BERTSCORE \citep{zhang2019bertscore}, YISI \citep{lo-2019-yisi,lo-larkin-2020-machine} and MOVERSCORE \citep{zhao2019moverscore}).

Recently, works such as PRISM \citep{thompson2020automatic} and BARTScore \citep{yuan2021bartscore} start to formulate evaluation as a generation task, which can not only make full use of pre-trained knowledge but also find more training data that can provide useful supervision for metrics to learn.

In this paper, we propose a framework that allows different types of signals to be incorporated into metrics. Concurrent with our work, UniEval \citep{zhong2022towards} formulates evaluation as a boolean question answering task and is trained on semi-supervised data. By contrast, our method is based on the generation formulation, which enables us to utilize a large amount of raw parallel data.

\section{Conclusions}
In this paper, we augment evaluation metrics with the ability to use different types of signal from data, which based on the assumption that a good evaluation metric not only should be informed of how to score different qualities of texts but also how high-quality texts are generated.
We achieve this goal by proposing a discriminative generation framework for the evaluation of generated texts, which outperform 8 existing top-performing metrics on 5 datasets from 19 languages.

\section{Limitations}
This study has potential limitations: (1) While different perspectives (e.g. informativeness, fluency, or factuality) of text could be evaluated, we assign one overall quality score, which cannot necessarily reflect the quality of a certain aspect. In the future work, more interpretable metrics could be designed specifically for a certain evaluation perspective. (2) Due to the contrastive training objective of \metricname, our metric may not be as good at predicting the absolute score of segments or systems compared to predicting their relative ranks. (3) We focus on evaluating automatically generated texts by machine systems. Human generated texts which might have different features from machine generated texts could be addressed in future work. (4) We study reference-based and source-based metrics separately. A combination of both could be studied in the future. 










 \bibliography{anthology,custom}


\appendix

\section{Appendix}
\label{sec:appendix}

\subsection{Training with Different Corpora}\label{sec:appendix-training-data}

We compare our model discriminatively trained on different training corpora: \metricname-*$^{20}$, \metricname-*$^{21}$ and \metricname-*$^{21}_{\text{mqm}}$ with baselines: COMET$^{20}$ which is trained on WMT DA corpus from 2017 to 2019 and COMET$^{21}_{\text{mqm}}$ which is trained on WMT DA corpus from 2017 to 2020 and further trained for 1 additional epoch on MQM-20.\footnote{COMET$^{20}$ uses the model wmt20-comet-da and COMET$^{21}_{\text{mqm}}$ uses the model wmt21-comet-mqm.} Tab.\ref{tab: MQM-data} shows that adding one more year's  DARR corpus (WMT-DA-20) as training data, \metricname-*$^{21}$ has better performance than \metricname-*$^{21}$. Besides, adding MQM in training has large performance improvement, although the total number of MQM training samples are much less than samples in DA corpus.
\begin{table}[!htp]\centering
\setlength\tabcolsep{2.4pt}
\caption{Segment Kendall’s Tau, system Pearson and system Kendall’s Tau of different metrics on MQM21 with different training data.The highest correlation for each language pair under each correlation method is bold.
}\label{tab: MQM-data}
\footnotesize
\begin{tabular}{lrrrrrrr}\toprule
&\multicolumn{3}{c}{en-de} &\multicolumn{3}{c}{zh-en} \\\cmidrule(lr){2-4}\cmidrule(lr){5-7}
&sys-p &sys-k &seg-k &sys-p &sys-k &seg-k \\\midrule
COMET$^{20}$ &82.3 &64.1 &18.4 &37.2 &41.0 &11.9 \\
COMET$^{21}_{\text{mqm}}$ &77.1 &66.7 &19.8 &55.9 &33.3 &\bftab14.1 \\\midrule\midrule
\metricname-B$^{20}$ &75.8 &53.8 &14.5 &42.5 &28.2 &10.0 \\
\metricname-B$^{21}$ &82.3 &56.4 &16.3 &41.6 &28.2 &10.9 \\
\metricname-B$^{21}_{\text{mqm}}$ &\bftab91.7 &\bftab79.5 &18.0 &56.5 &46.2 &11.5 \\\midrule
\metricname-L$^{20}$ &79.4 &56.4 &15.8 &44.1 &38.5 &11.3 \\
\metricname-L$^{21}$ &83.9 &59.0 &17.7 &44.0 &33.3 &11.1 \\
\metricname-L$^{21}_{\text{mqm}}$ &88.9 &74.4 &\bftab20.7 &58.9 &48.7 &13.6 \\\midrule
\metricname-XL$^{21}$ &83.5 &61.5 &18.0 &47.3 &35.9 &12.4 \\
\metricname-XL$^{20}$ &78.7 &53.8 &16.6 &53.4 &43.6 &12.2 \\
\metricname-XL$^{21}_{\text{mqm}}$ &86.0 &69.2 &19.2 &\bftab62.9 &\bftab53.8 &12.5 \\
\bottomrule
\end{tabular}
\end{table}

\subsection{Training Details}\label{sec:appendix-hyperparameters}
For generative training, our model is trained on ParaCotta, and WMT-19 is used as validation set, while WMT-20 is used as test set. The hyper-parameter tuned is learning rate. For discriminative training, on the base of the generative model, we further train our model on the z-score of WMT DA corpus from 2017 to 2019, using the discriminative loss function in equation\ref{equ:L_sup}. The hyper-parameters tuned are learning rate, dropout rate and $\alpha$. We choose 4 language pairs (ru-en, en-ru, en-pl, en-cs) as the validation set to tune the hyper-parameters and pick the best model checkpoint. 

Tab.\ref{tab: hyperparameters} shows the hyper-parameters, training time and computing resources of \metricname. We find as the model size increases, we need less training steps to get the best performance on the validation set, so maximum training steps is decreased for larger models. We use the linear learning rate scheduler and use 10\% of the maximum training steps as the wamp up step. In the table, when GPU is larger than 1, we use model parallelism. 

\begin{table*}[!htp]\centering
\setlength\tabcolsep{5.2pt}
\caption{Hyper-parameters, training time and computing resources. Max-step means the maximum training steps. Save-step means we save a checkpoint every save-step. GPU shows the GPU type and the number of GPUs. A6000 is the NVIDIA $\text{RTX}^{\text{TM}}$ A6000. Time is the wall clock training time.}\label{tab: hyperparameters}
\footnotesize
\begin{tabular}{lrrrrrrrrrrr}\toprule
&max-step &save-step &batch-size &warm-up &learning-rate &dropout &$\alpha$ &GPU &time \\\midrule
\metricname-B$_{\text{un}}$ &10,000 &500 &10  &1,000 &5.00e-5 &0.10 &- &A6000*1 &0.6h \\
\metricname-L$_{\text{un}}$ &10,000 &500 &10  &1,000 &5.00e-5 &0.10 &- &A6000*1 &1h \\
\metricname-XL$_{\text{un}}$ &10,000 &500 &10  &1,000 &5.00e-5 &0.10 &- &A6000*1 &3h \\
\metricname-B$_{\text{sup}}$ &300,000 &10,000 &32  &0 &1.00e-5 &0.05 &1 &A6000*1 &60h \\
\metricname-L$_{\text{sup}}$ &200,000 &10,000 &32  &0 &1.00e-5 &0.05 &1 &A6000*2 &80h \\
\metricname-XL$_{\text{sup}}$ &50,000 &5,000 &32  &0 &1.00e-5 &0.05 &1 &A6000*4 &40h \\
\bottomrule
\end{tabular}
\end{table*}

\subsection{WMT-DA system level results}\label{sec:appendix-WMT-DA-sys}
Besides segment level correlation, we also calculate system level correlation on WMT DA20 corpus, and the results are shown in Tab.\ref{tab: wmt-da-sys-p-x-en},\ref{tab: wmt-da-sys-p-en-x},\ref{tab: wmt-da-sys-k-x-en},\ref{tab: wmt-da-sys-k-en-x}.

\begin{table*}[!htp]\centering
\caption{System level Pearson correlations on language pairs with English as target for the WMT DA20 corpus. Avg. denotes the average correlation across all x-en language pairs.}\label{tab: wmt-da-sys-p-x-en}
\footnotesize
\begin{tabular}{lrrrrrrrrrrrr}\toprule
&cs-en &de-en &iu-en &ja-en &km-en &pl-en &ps-en &ru-en &ta-en &zh-en &Avg \\\midrule
\multicolumn{12}{l}{UNSUPERVISED METHODS}  \\\midrule
sentBLEU &84.4 &97.8 &65.2 &97.4 &96.9 &50.1 &88.8 &91.7 &92.5 &94.8 &85.9 \\
BERTScore &81.6 &99.8 &73.0 &97.4 &95.0 &59.2 &93.0 &92.7 &89.5 &96.1 &87.7 \\
PRISM &81.8 &99.8 &83.3 &97.4 &95.0 &50.2 &96.6 &90.8 &89.8 &95.7 &88.0 \\
BARTScore &84.4 &99.8 &79.2 &97.7 &94.4 &52.7 &95.1 &92.4 &91.4 &95.8 &88.3 \\\midrule
\metricname-B$_{\text{un}}$ &86.1 &99.9 &80.9 &98.1 &95.2 &52.9 &95.8 &93.1 &90.7 &96.1 &88.9 \\
\metricname-L$_{\text{un}}$ &84.4 &99.9 &83.8 &98.1 &95.2 &53.6 &96.7 &92.9 &89.9 &96.2 &89.1 \\
\metricname-XL$_{\text{un}}$ &83.9 &99.9 &84.0 &97.8 &95.0 &53.7 &96.7 &92.5 &89.4 &96.2 &88.9 \\\midrule\midrule
\multicolumn{12}{l}{SUPERVISED METHODS}  \\\midrule
bleurt &78.9 &99.7 &84.0 &96.5 &99.4 &56.6 &95.7 &89.8 &92.0 &94.7 &88.8 \\
comet &78.3 &99.8 &85.2 &96.4 &97.1 &59.1 &94.1 &92.3 &88.0 &95.2 &88.6 \\\midrule
\metricname-B$_{\text{sup}}$ &81.9 &99.4 &78.9 &98.2 &97.8 &57.4 &95.1 &91.9 &88.1 &96.4 &88.5 \\
\metricname-L$_{\text{sup}}$ &80.4 &99.3 &77.8 &98.0 &98.6 &59.4 &95.4 &92.2 &89.6 &96.6 &88.7 \\
\metricname-XL$_{\text{sup}}$ &79.3 &99.3 &81.7 &97.4 &98.3 &58.0 &95.9 &92.0 &88.4 &95.5 &88.6 \\
\bottomrule
\end{tabular}
\end{table*}

\begin{table*}[!htp]\centering
\caption{System level Pearson correlations on language pairs with English as source for the WMT DA20 corpus. Avg. denotes the average correlation across all en-x language pairs and Avg-all denotes the average correlation across all language pairs.}\label{tab: wmt-da-sys-p-en-x}
\footnotesize
\begin{tabular}{lrrrrrrrrrrr}\toprule
&en-cs &en-de &en-iu &en-ja &en-pl &en-ru &en-ta &en-zh &Avg &Avg-all \\\midrule
\multicolumn{11}{l}{UNSUPERVISED METHODS}  \\\midrule
sentBLEU &84.1 &93.4 &13.5 &94.6 &95.0 &98.1 &88.2 &92.8 &82.5 &84.4 \\
BERTScore &88.4 &95.5 &68.4 &97.0 &91.4 &97.2 &95.4 &92.4 &90.7 &89.1 \\
PRISM &94.9 &95.8 &85.9 &93.2 &95.8 &72.4 &91.6 &96.8 &90.8 &89.3 \\
BARTScore &90.8 &94.5 &86.2 &94.7 &95.5 &61.8 &95.8 &94.7 &89.3 &88.7 \\\midrule
\metricname-B$_{\text{un}}$ &89.1 &95.9 &65.8 &95.5 &96.0 &63.8 &96.1 &96.4 &87.3 &88.2 \\
\metricname-L$_{\text{un}}$ &91.0 &95.9 &67.0 &95.9 &96.1 &63.0 &96.6 &96.6 &87.8 &88.5 \\
\metricname-XL$_{\text{un}}$ &92.4 &95.2 &63.2 &94.9 &96.4 &70.6 &96.7 &97.0 &88.3 &88.6 \\\midrule\midrule
\multicolumn{11}{l}{SUPERVISED METHODS}  \\\midrule
bleurt &98.8 &95.4 &74.8 &95.8 &98.3 &82.6 &92.6 &98.3 &92.1 &90.2 \\
comet &97.8 &97.2 &86.0 &97.4 &98.1 &92.5 &94.4 &98.2 &95.2 &91.5 \\\midrule
\metricname-B$_{\text{sup}}$ &93.7 &97.2 &67.9 &98.2 &97.8 &92.9 &93.7 &97.6 &92.4 &90.2 \\
\metricname-L$_{\text{sup}}$ &96.4 &97.3 &67.4 &97.4 &98.0 &94.8 &95.9 &97.8 &93.1 &90.7 \\
\metricname-XL$_{\text{sup}}$ &96.8 &96.8 &61.3 &95.9 &98.3 &95.7 &95.0 &98.6 &92.3 &90.2 \\
\bottomrule
\end{tabular}
\end{table*}

\begin{table*}[!htp]\centering
\caption{System level Kendall’s Tau correlations on language pairs with English as target for the WMT DA20 corpus. Avg. denotes the average correlation across all x-en language pairs.}\label{tab: wmt-da-sys-k-x-en}
\footnotesize
\begin{tabular}{lrrrrrrrrrrrr}\toprule
&cs-en &de-en &iu-en &ja-en &km-en &pl-en &ps-en &ru-en &ta-en &zh-en &Avg \\\midrule
\multicolumn{12}{l}{UNSUPERVISED METHODS}  \\\midrule
sentBLEU &78.8 &75.8 &45.5 &73.3 &61.9 &27.5 &60.0 &60.0 &69.2 &85.0 &63.7 \\
BERTScore &78.8 &72.7 &63.6 &73.3 &71.4 &47.3 &86.7 &52.7 &67.0 &80.0 &69.4 \\
PRISM &75.8 &72.7 &67.3 &86.7 &71.4 &34.1 &86.7 &56.4 &64.8 &80.0 &69.6 \\
BARTScore &81.8 &75.8 &67.3 &86.7 &71.4 &36.3 &86.7 &56.4 &64.8 &83.3 &71.0 \\\midrule
\metricname-B$_{\text{un}}$ &84.8 &72.7 &70.9 &82.2 &71.4 &36.3 &86.7 &56.4 &67.0 &81.7 &71.0 \\
\metricname-L$_{\text{un}}$ &81.8 &75.8 &74.5 &86.7 &71.4 &36.3 &86.7 &56.4 &64.8 &80.0 &71.4 \\
\metricname-XL$_{\text{un}}$ &78.8 &75.8 &74.5 &82.2 &71.4 &36.3 &86.7 &56.4 &62.6 &80.0 &70.5 \\\midrule\midrule
\multicolumn{12}{l}{SUPERVISED METHODS}  \\\midrule
bleurt &75.8 &81.8 &63.6 &77.8 &100.0 &42.9 &86.7 &49.1 &60.4 &73.3 &71.1 \\
comet &72.7 &75.8 &63.6 &77.8 &100.0 &40.7 &86.7 &56.4 &62.6 &73.3 &71.0 \\\midrule
\metricname-B$_{\text{sup}}$ &72.7 &75.8 &56.4 &82.2 &90.5 &42.9 &86.7 &56.4 &60.4 &83.3 &70.7 \\
\metricname-L$_{\text{sup}}$ &75.8 &78.8 &56.4 &77.8 &100.0 &42.9 &86.7 &52.7 &60.4 &71.7 &70.3 \\
\metricname-XL$_{\text{sup}}$ &78.8 &78.8 &67.3 &82.2 &90.5 &45.1 &86.7 &52.7 &62.6 &76.7 &72.1 \\
\bottomrule
\end{tabular}
\end{table*}

\begin{table*}[!htp]\centering
\caption{System level Kendall’s Tau correlations on language pairs with English as source for the WMT DA20 corpus. Avg. denotes the average correlation across all en-x language pairs and Avg-all denotes the average correlation across all language pairs.}\label{tab: wmt-da-sys-k-en-x}
\footnotesize
\begin{tabular}{lrrrrrrrrrrr}\toprule
&en-cs &en-de &en-iu &en-ja &en-pl &en-ru &en-ta &en-zh &Avg &Avg-all \\\midrule
\multicolumn{11}{l}{UNSUPERVISED METHODS}  \\\midrule
sentBLEU &51.5 &80.2 &23.6 &85.5 &60.4 &94.4 &86.7 &72.7 &69.4 &66.2 \\
BERTScore &51.5 &80.2 &34.5 &85.5 &56.0 &94.4 &84.8 &72.7 &70.0 &69.6 \\
PRISM &81.8 &86.8 &45.5 &81.8 &67.0 &61.1 &73.3 &81.8 &72.4 &70.8 \\
BARTScore &54.5 &82.4 &41.8 &78.2 &64.8 &61.1 &84.8 &75.8 &67.9 &69.7 \\\midrule
\metricname-B$_{\text{un}}$ &54.5 &84.6 &34.5 &85.5 &67.0 &61.1 &86.7 &69.7 &68.0 &69.7 \\
\metricname-L$_{\text{un}}$ &51.5 &82.4 &34.5 &89.1 &67.0 &61.1 &84.8 &75.8 &68.3 &70.0 \\
\metricname-XL$_{\text{un}}$ &51.5 &82.4 &27.3 &85.5 &67.0 &66.7 &84.8 &72.7 &67.2 &69.0 \\\midrule\midrule
\multicolumn{11}{l}{SUPERVISED METHODS}  \\\midrule
bleurt &90.9 &84.6 &34.5 &81.8 &73.6 &66.7 &73.3 &84.8 &73.8 &72.3 \\
comet &90.9 &84.6 &38.2 &74.5 &73.6 &72.2 &77.1 &81.8 &74.1 &72.4 \\\midrule
\metricname-B$_{\text{sup}}$ &75.8 &86.8 &34.5 &85.5 &69.2 &72.2 &77.1 &84.8 &73.3 &71.8 \\
\metricname-L$_{\text{sup}}$ &90.9 &89.0 &30.9 &85.5 &69.2 &72.2 &79.0 &84.8 &75.2 &72.5 \\
\metricname-XL$_{\text{sup}}$ &90.9 &89.0 &23.6 &89.1 &71.4 &72.2 &77.1 &87.9 &75.2 &73.5 \\
\bottomrule
\end{tabular}
\end{table*}

\subsection{Unsupvised \metricname-XXL Results}\label{sec:appendix-WMT-DA-mt5-xxl}
In Tab.\ref{tab: mT5-XXl}, we show results on WMT DA20 corpus using generative \metricname-XXL.
\begin{table*}[!htp]\centering
\caption{Segment Kendall’s Tau, system Pearson and system Kendall’s Tau correlations on all language pairs for the WMT DA20 corpus using generative \metricname-XXL. Avg-en denotes the average correlation across all x-en language pairs, Avg-x denotes the average correlation across all en-x language pairs, and Avg denotes the average correlation across all language pairs.}\label{tab: mT5-XXl}
\footnotesize
\begin{tabular}{lrrrrrrrrrrrr}\toprule
&cs-en &de-en &iu-en &ja-en &km-en &pl-en &ps-en &ru-en &ta-en &zh-en &Avg-en \\\midrule
seg-k &13.5 &49.0 &26.9 &27.8 &29.9 &8.7 &16.7 &13.9 &24.4 &15.8 &22.7 \\
sys-p &83.0 &99.9 &83.7 &97.6 &95.2 &55.2 &96.9 &92.4 &90.0 &96.2 &89.0 \\
sys-k &72.7 &75.8 &74.5 &77.8 &71.4 &36.3 &86.7 &56.4 &64.8 &80.0 &69.6 \\\toprule
&en-cs &en-de &en-iu &en-ja &en-pl &en-ru &en-ta &en-zh &Avg-x &Avg & \\\midrule
seg-k &60.4 &43.8 &33.8 &60.6 &40.0 &26.9 &65.3 &47.2 &47.3 &33.6 & \\
sys-p &93.1 &95.6 &64.2 &96.1 &96.4 &68.5 &96.7 &97.2 &88.5 &88.8 & \\
sys-k &60.6 &82.4 &27.3 &89.1 &64.8 &66.7 &82.9 &75.8 &68.7 &69.2 & \\
\bottomrule
\end{tabular}
\end{table*}

\subsection{WMT-DA Source Based Evaluation Results}\label{sec:appendix-WMT-DA-src-based}
We conduct source-based evaluation experiments on WMT DA20 corpus using supervised \metricname. We compare our supervised source-based model with COMET$_{\text{src}}$  baseline \citep{rei2021references}, which is a source-based metric trained to predict WMT DA scores from 2017 to 2019.\footnote{COMET$_{\text{src}}$ uses the model wmt20-comet-qe-da.} Tab.\ref{tab: seg-src} illustrates the segment Kendall’s Tau correlation of diverse evaluation metrics on WMT-DA. The results show that at segment level \metricname-B$_{\text{src}}$ is comparable to the baseline and larger models surpass the baseline. Our system level results, shown in Tab.\ref{tab: source-sys-p},\ref{tab: source-sys-k} are also comparable to or better than the baseline for most language pairs except iu-en and en-iu. In all tables, Avg-en denotes the average correlation across all x-en language pairs; Avg-x denotes the average correlation for all en-x language pairs; Avg denotes the average correlation for all language pairs.

\begin{table*}[!htp]\centering
\caption{Segment level Kendall’s Tau correlations on WMT DA20 corpus using source-based models. The highest correlation for each language pair is bold.}\label{tab: seg-src}
\footnotesize
\begin{tabular}{lrrrrrrrrrrrr}\toprule
&cs-en &de-en &iu-en &ja-en &km-en &pl-en &ps-en &ru-en &ta-en &zh-en &Avg-en \\\midrule
COMET$_{\text{src}}$ &\bftab9.2 &40.8 &\bftab3.2 &15.3 &14.9 &\bftab4.6 &9.2 &\bftab10.1 &16.7 &9.2 &13.3 \\
\metricname-B$_{\text{src}}$ &7.5 &\bftab43.2 &0.7 &16.2 &11.9 &4.4 &9.7 &7.5 &19.2 &\bftab9.4 &12.9 \\
\metricname-L$_{\text{src}}$ &5.7 &35.9 &0.7 &19.2 &16.7 &4.1 &10.5 &8.4 &22.5 &8.1 &13.2 \\
\metricname-XL$_{\text{src}}$ &7.1 &41.7 &2.6 &\bftab22.1 &\bftab22.5 &3.0 &\bftab10.6 &7.3 &\bftab24.7 &7.4 &\bftab14.9 \\\toprule
&en-cs &en-de &en-iu &en-ja &en-pl &en-ru &en-ta &en-zh &Avg-x &Avg & \\\midrule
COMET$_{\text{src}}$ &61.3 &34.6 &-6.3 &46.7 &35.8 &26.4 &51.2 &39.8 &36.2 &23.5 & \\
\metricname-B$_{\text{src}}$ &55.3 &36.4 &\bftab0.0 &51.2 &28.6 &24.7 &57.1 &41.2 &36.8 &23.5 & \\
\metricname-L$_{\text{src}}$ &60.5 &37.1 &-5.2 &56.9 &37.4 &27.2 &63.9 &44.3 &40.3 &25.2 & \\
\metricname-XL$_{\text{src}}$ &\bftab62.7 &\bftab40.0 &-2.6 &\bftab58.7 &\bftab37.6 &\bftab28.2 &\bftab66.2 &\bftab45.6 &\bftab42.0 &\bftab27.0 & \\
\bottomrule
\end{tabular}
\end{table*}

\begin{table*}[!htp]\centering
\caption{System level Pearson correlations on WMT DA20 corpus using source based models. }\label{tab: source-sys-p}
\footnotesize
\begin{tabular}{lrrrrrrrrrrrr}\toprule
&cs-en &de-en &iu-en &ja-en &km-en &pl-en &ps-en &ru-en &ta-en &zh-en &Avg-en \\\midrule
COMET$_{\text{src}}$ &75.5 &93.9 &\bftab70.6 &89.2 &89.6 &44.8 &83.2 &88.3 &79.5 &84.7 &\bftab79.9 \\
\metricname-B$_{\text{src}}$ &\bftab82.6 &\bftab99.7 &-7.8 &94.1 &86.9 &47.4 &92.2 &90.0 &81.6 &\bftab95.1 &76.2 \\
\metricname-L$_{\text{src}}$&73.9 &\bftab99.7 &-17.3 &\bftab96.2 &92.2 &\bftab52.6 &92.0 &91.1 &85.5 &92.6 &75.8 \\
\metricname-XL$_{\text{src}}$ &73.0 &99.5 &1.0 &95.9 &\bftab99.1 &51.3 &\bftab94.0 &\bftab92.0 &\bftab87.9 &91.5 &78.5 \\\toprule
&en-cs &en-de &en-iu &en-ja &en-pl &en-ru &en-ta &en-zh &Avg-x &Avg & \\\midrule
COMET$_{\text{src}}$ &\bftab98.9 &90.1 &\bftab86.3 &95.2 &\bftab96.9 &80.0 &88.8 &\bftab97.5 &\bftab91.7 &\bftab85.2 & \\
\metricname-B$_{\text{src}}$ &95.8 &\bftab97.3 &31.5 &\bftab96.8 &93.5 &79.4 &91.7 &93.7 &85.0 &80.1 & \\
\metricname-L$_{\text{src}}$ &98.2 &96.6 &27.4 &\bftab96.8 &96.2 &84.9 &94.9 &95.7 &86.3 &80.5 & \\
\metricname-XL$_{\text{src}}$ &98.5 &96.3 &22.4 &96.4 &96.8 &\bftab88.7 &\bftab95.2 &96.0 &86.3 &82.0 & \\
\bottomrule
\end{tabular}
\end{table*}

\begin{table*}[!htp]\centering
\caption{System level Kendall’s Tau correlations on WMT DA20 corpus using source based models.}\label{tab: source-sys-k}
\footnotesize
\begin{tabular}{lrrrrrrrrrrrr}\toprule
&cs-en &de-en &iu-en &ja-en &km-en &pl-en &ps-en &ru-en &ta-en &zh-en &Avg-en \\\midrule
COMET$_{\text{src}}$ &69.7 &\bftab78.8 &\bftab52.7 &77.8 &\bftab90.5 &29.7 &73.3 &45.5 &51.6 &55.0 &\bftab62.5 \\
\metricname-B$_{\text{src}}$ &66.7 &66.7 &-9.1 &64.4 &61.9 &\bftab31.9 &\bftab86.7 &\bftab56.4 &60.4 &65.0 &55.1 \\
\metricname-L$_{\text{src}}$ &\bftab72.7 &\bftab78.8 &-12.7 &82.2 &81.0 &\bftab31.9 &73.3 &52.7 &56.0 &\bftab68.3 &58.4 \\
\metricname-XL$_{\text{src}}$ &\bftab72.7 &\bftab78.8 &5.5 &\bftab86.7 &\bftab90.5 &29.7 &73.3 &52.7 &\bftab64.8 &\bftab68.3 &62.3 \\\toprule
&en-cs &en-de &en-iu &en-ja &en-pl &en-ru &en-ta &en-zh &Avg-x &Avg & \\\midrule
COMET$_{\text{src}}$ &84.8 &\bftab80.2 &\bftab60.0 &70.9 &\bftab80.2 &\bftab66.7 &56.2 &\bftab84.8 &\bftab73.0 &\bftab67.1 & \\
\metricname-B$_{\text{src}}$ &84.8 &78.0 &12.7 &\bftab81.8 &71.4 &61.1 &65.7 &81.8 &67.2 &60.5 & \\
\metricname-L$_{\text{src}}$ &\bftab93.9 &78.0 &1.8 &78.2 &71.4 &\bftab66.7 &77.1 &81.8 &68.6 &63.0 & \\
\metricname-XL$_{\text{src}}$ &\bftab93.9 &\bftab80.2 &1.8 &78.2 &67.0 &\bftab66.7 &\bftab82.9 &78.8 &68.7 &65.1 & \\
\bottomrule
\end{tabular}
\end{table*}

\subsection{Effectiveness of Generative Training}\label{sec:appendix-WMT-DA-ablation}
To show the importance of unsupervised generative training, we compare the performance of \metricname-B and \metricname-L with and without unsupervised generatively training in Tab.\ref{tab: without-mT5-B},\ref{tab: without-mT5-L},\ref{tab: with-mT5-B},\ref{tab: with-mT5-L}. Avg denotes the average correlation for all language pairs.

\begin{table*}[!htp]\centering
\caption{Segment Kendall’s Tau, system Pearson and system Kendall’s Tau correlations for WMT DA20 corpus without unsupervised training using \metricname-B.}\label{tab: without-mT5-B}
\footnotesize
\begin{tabular}{lrrrrrrrrrrr}\toprule
 &cs-en &de-en &iu-en &ja-en &km-en &pl-en &ps-en &ru-en &ta-en &zh-en \\\midrule
seg-k &14.0 &48.4 &27.8 &28.8 &30.7 &10.2 &18.0 &13.1 &24.4 &16.4 \\
sys-p &82.1 &99.2 &79.9 &97.8 &97.7 &56.3 &94.7 &91.8 &88.5 &96.3 \\
sys-k &72.7 &75.8 &56.4 &82.2 &90.5 &40.7 &86.7 &56.4 &62.6 &80.0 \\\midrule
 &en-cs &en-de &en-iu &en-ja &en-pl &en-ru &en-ta &en-zh &Avg & \\
seg-k &60.9 &43.4 &33.4 &59.3 &39.6 &29.3 &63.7 &48.6 &33.9 & \\
sys-p &93.9 &96.7 &64.0 &97.2 &97.9 &91.5 &93.9 &96.9 &89.8 & \\
sys-k &69.7 &86.8 &30.9 &85.5 &69.2 &72.2 &77.1 &81.8 &71.0 & \\
\bottomrule
\end{tabular}
\end{table*}

\begin{table*}[!htp]\centering
\caption{Segment Kendall’s Tau, system Pearson and system Kendall’s Tau correlations for WMT DA20 corpus without unsupervised training  using model \metricname-L.}\label{tab: without-mT5-L}
\footnotesize
\begin{tabular}{lrrrrrrrrrrr}\toprule
 &cs-en &de-en &iu-en &ja-en &km-en &pl-en &ps-en &ru-en &ta-en &zh-en \\\midrule
seg-k &13.2 &48.8 &30.5 &28.9 &30.7 &9.6 &18.4 &14.5 &25.6 &16.1 \\
sys-p &81.5 &99.3 &79.8 &97.9 &97.9 &58.2 &95.2 &92.2 &89.0 &96.5 \\
sys-k &72.7 &78.8 &56.4 &77.8 &90.5 &40.7 &86.7 &60.0 &60.4 &73.3 \\\midrule
 &en-cs &en-de &en-iu &en-ja &en-pl &en-ru &en-ta &en-zh &Avg & \\
seg-k &65.1 &44.5 &14.8 &63.0 &44.6 &30.5 &67.2 &51.2 &34.3 & \\
sys-p &96.1 &96.9 &54.4 &96.8 &98.1 &93.1 &95.6 &97.2 &89.8 & \\
sys-k &90.9 &86.8 &20.0 &85.5 &71.4 &72.2 &79.0 &87.9 &71.7 & \\
\bottomrule
\end{tabular}
\end{table*}

\begin{table*}[!htp]\centering
\caption{Segment Kendall’s Tau, system Pearson and system Kendall’s Tau correlations for WMT DA20 corpus with unsupervised training  using model \metricname-B.}\label{tab: with-mT5-B}
\footnotesize
\begin{tabular}{lrrrrrrrrrrr}\toprule
 &cs-en &de-en &iu-en &ja-en &km-en &pl-en &ps-en &ru-en &ta-en &zh-en \\\midrule
seg-k &13.9 &48.5 &29.2 &28.1 &30.3 &9.6 &17.4 &13.1 &23.9 &15.5 \\
sys-p &81.9 &99.4 &78.9 &98.2 &97.8 &57.4 &95.1 &91.9 &88.1 &96.4 \\
sys-k &72.7 &75.8 &56.4 &82.2 &90.5 &42.9 &86.7 &56.4 &60.4 &83.3 \\\midrule
 &en-cs &en-de &en-iu &en-ja &en-pl &en-ru &en-ta &en-zh &Avg & \\
seg-k &61.2 &43.1 &32.4 &60.5 &38.9 &28.9 &64.1 &48.5 &33.7 & \\
sys-p &93.7 &97.2 &67.9 &98.2 &97.8 &92.9 &93.7 &97.6 &90.2 & \\
sys-k &75.8 &86.8 &34.5 &85.5 &69.2 &72.2 &77.1 &84.8 &71.8 & \\
\bottomrule
\end{tabular}
\end{table*}

\begin{table*}[!htp]\centering
\caption{Segment Kendall’s Tau, system Pearson and system Kendall’s Tau correlations for WMT DA20 corpus with unsupervised training using \metricname-L.}\label{tab: with-mT5-L}
\footnotesize
\begin{tabular}{lrrrrrrrrrrr}\toprule
 &cs-en &de-en &iu-en &ja-en &km-en &pl-en &ps-en &ru-en &ta-en &zh-en \\\midrule
seg-k &14.0 &49.3 &28.5 &28.9 &30.1 &8.3 &17.6 &15.3 &25.9 &16.3 \\
sys-p &80.4 &99.3 &77.8 &98.0 &98.6 &59.4 &95.4 &92.2 &89.6 &96.6 \\
sys-k &75.8 &78.8 &56.4 &77.8 &100.0 &42.9 &86.7 &52.7 &60.4 &71.7 \\\midrule
 &en-cs &en-de &en-iu &en-ja &en-pl &en-ru &en-ta &en-zh &Avg & \\
seg-k &66.6 &45.6 &33.4 &62.8 &43.9 &31.8 &66.7 &52.2 &35.4 & \\
sys-p &96.4 &97.3 &67.4 &97.4 &98.0 &94.8 &95.9 &97.8 &90.7 & \\
sys-k &90.9 &89.0 &30.9 &85.5 &69.2 &72.2 &79.0 &84.8 &72.5 & \\
\bottomrule
\end{tabular}
\end{table*}

\subsection{MultiSumm Segment Level Pearson Correlation Result}\label{sec:appendix-multi-sum}
Tab.\ref{tab: MultiSumm}/Tab.\ref{tab: MultiSumm-Pearson} illustrates the segment Kendall’s Tau/Pearson correlation of diverse evaluation methods for 8 language pairs on MultiSumm dataset.

\begin{table*}[!htp]\centering
\setlength\tabcolsep{2.4pt}
\caption{Segment level Kendall’s Tau correlation on MultiSumm
corpus. The highest correlation is bold}\label{tab: MultiSumm}
\footnotesize
\begin{tabular}{lrrrrrrrrrrrrrrrrrrr}\toprule
&\multicolumn{9}{c}{Focus} &\multicolumn{9}{c}{Coverage} \\\cmidrule(lr){2-10}\cmidrule(lr){11-19}
&EN &ID &FR &TR &ZH &RU &DE &ES &AVG &EN &ID &FR &TR &ZH &RU &DE &ES &AVG \\\midrule
ROUGE1 &\bftab49.0 &48.9 &48.1 &39.6 &48.5 &7.5 &47.1 &36.8 &40.7 &52.9 &50.5 &48.5 &32.0 &47.5 &9.8 &8.9 &51.6 &37.7 \\
ROUGE2 &34.7 &51.1 &36.5 &35.4 &50.5 &-67.7 &35.6 &28.4 &25.6 &52.9 &41.9 &34.0 &30.0 &43.4 &-66.7 &1.3 &20.9 &19.7 \\
ROUGEL &44.9 &46.7 &48.1 &37.5 &48.5 &5.4 &49.4 &38.9 &39.9 &54.9 &46.2 &44.3 &30.0 &49.5 &7.8 &11.4 &42.9 &35.9 \\
COMET &36.7 &55.6 &44.2 &33.3 &48.5 &39.8 &60.9 &43.2 &45.3 &41.2 &48.4 &25.8 &32.0 &47.5 &29.4 &31.6 &51.6 &38.4 \\
BERTScore &34.7 &51.1 &48.1 &\bftab50.0 &36.6 &7.5 &63.2 &43.2 &41.8 &49.0 &52.7 &44.3 &\bftab38.0 &41.4 &-29.4 &31.6 &49.5 &34.6 \\
BLEURT &40.8 &51.1 &55.8 &43.8 &50.5 &57.0 &47.1 &\bftab49.5 &49.4 &45.1 &46.2 &42.3 &34.0 &45.5 &\bftab70.6 &21.5 &60.4 &45.7 \\
PRISM &32.7 &48.9 &57.7 &25.0 &46.5 &\bftab61.3 &54.0 &43.2 &46.2 &45.1 &59.1 &34.0 &\bftab38.0 &\bftab59.6 &54.9 &31.6 &53.8 &47.0 \\
BARTScore &32.7 &55.6 &63.5 &41.7 &54.5 &41.9 &\bftab67.8 &\bftab49.5 &50.9 &41.2 &44.1 &44.3 &36.0 &49.5 &52.9 &\bftab34.2 &60.4 &45.3 \\\midrule
\metricname-B$_{\text{un}}$ &42.9 &51.1 &\bftab69.2 &43.8 &48.5 &52.7 &\bftab67.8 &43.2 &52.4 &51.0 &52.7 &44.3 &30.0 &47.5 &52.9 &26.6 &60.4 &45.7 \\
\metricname-L$_{\text{un}}$ &26.5 &\bftab60.0 &67.3 &39.6 &\bftab56.4 &\bftab61.3 &63.2 &45.3 &\bftab52.5 &33.3 &59.1 &\bftab52.6 &28.0 &\bftab59.6 &64.7 &21.5 &60.4 &\bftab47.4 \\
\metricname-XL$_{\text{un}}$ &24.5 &55.6 &59.6 &39.6 &54.5 &57.0 &65.5 &\bftab49.5 &50.7 &33.3 &63.4 &44.3 &26.0 &49.5 &68.6 &19.0 &\bftab67.0 &46.4 \\
\metricname-B$_{\text{sup}}$ &46.9 &44.4 &38.5 &35.4 &42.6 &-3.2 &54.0 &30.5 &36.1 &\bftab58.8 &41.9 &32.0 &16.0 &39.4 &-37.3 &16.5 &38.5 &25.7 \\
\metricname-L$_{\text{sup}}$ &38.8 &\bftab60.0 &42.3 &37.5 &42.6 &-14.0 &42.5 &32.6 &35.3 &54.9 &52.7 &32.0 &26.0 &37.4 &-41.2 &13.9 &42.9 &27.3 \\
\metricname-XL$_{\text{sup}}$ &24.5 &\bftab60.0 &42.3 &\bftab50.0 &48.5 &14.0 &65.5 &41.1 &43.2 &52.9 &\bftab61.3 &15.5 &32.0 &37.4 &-7.8 &19.0 &49.5 &32.5 \\
\bottomrule
\end{tabular}
\end{table*}

\begin{table*}[!htp]\centering
\setlength\tabcolsep{2.4pt}
\caption{Segment level Pearson correlation on MultiSumm
corpus.}\label{tab: MultiSumm-Pearson}
\footnotesize
\begin{tabular}{lrrrrrrrrrrrrrrrrrrr}\toprule
&\multicolumn{9}{c}{Focus} &\multicolumn{9}{c}{Coverage} \\\cmidrule(lr){2-10}\cmidrule(lr){11-19}
&EN &ID &FR &TR &ZH &RU &DE &ES &AVG &EN &ID &FR &TR &ZH &RU &DE &ES &AVG \\\midrule
ROUGE1 &59.0 &70.2 &68.7 &81.0 &82.6 &52.0 &87.3 &60.1 &70.1 &62.3 &70.5 &66.2 &75.6 &77.9 &46.6 &88.8 &63.7 &68.7 \\
ROUGE3 &53.6 &62.8 &68.0 &77.8 &78.7 &51.6 &85.8 &61.2 &68.1 &53.5 &65.1 &67.3 &73.6 &73.6 &47.2 &88.2 &64.9 &66.9 \\
ROUGEL &58.2 &69.3 &68.8 &80.0 &81.6 &51.3 &86.8 &61.0 &69.7 &61.0 &70.7 &66.3 &75.5 &78.1 &46.2 &88.3 &64.7 &68.6 \\
COMET &50.6 &67.6 &53.1 &72.6 &73.1 &49.9 &84.8 &55.6 &62.8 &50.1 &68.7 &50.2 &76.6 &70.8 &49.9 &80.1 &62.1 &62.8 \\
BERTScore &58.5 &69.8 &71.8 &83.2 &77.7 &49.4 &89.6 &58.0 &69.7 &61.7 &71.6 &70.2 &80.0 &75.9 &39.0 &89.4 &66.4 &68.9 \\
BLEURT &52.4 &61.1 &70.5 &79.0 &72.6 &55.8 &87.4 &66.0 &69.1 &59.5 &61.7 &71.4 &79.0 &75.3 &58.7 &87.8 &65.2 &71.0 \\
PRISM &59.3 &59.5 &68.1 &78.9 &70.8 &42.5 &87.0 &53.9 &65.8 &61.2 &60.9 &64.4 &78.3 &69.3 &45.5 &89.0 &60.0 &66.8 \\
BARTScore &61.1 &69.0 &72.0 &80.3 &75.6 &54.6 &89.3 &62.5 &70.8 &59.9 &67.7 &68.9 &80.1 &70.9 &57.7 &90.7 &66.3 &70.7 \\\midrule
\metricname-B$_{\text{un}}$ &59.0 &64.9 &70.8 &79.6 &75.6 &50.0 &88.7 &58.5 &68.9 &59.0 &63.9 &68.2 &79.4 &71.5 &54.9 &90.1 &65.7 &69.8 \\
\metricname-L$_{\text{un}}$ &58.5 &68.6 &71.1 &80.6 &76.4 &49.3 &89.1 &59.1 &69.2 &58.2 &68.5 &70.1 &81.4 &73.0 &55.8 &90.4 &66.4 &70.8 \\
\metricname-XL$_{\text{un}}$ &57.8 &69.9 &70.2 &80.2 &77.0 &50.2 &88.9 &59.7 &69.1 &58.0 &70.4 &70.2 &81.3 &72.9 &54.7 &90.0 &65.7 &70.4 \\
\metricname-B$_{\text{sup}}$ &55.7 &63.5 &58.9 &70.6 &71.9 &26.6 &84.4 &42.8 &58.7 &51.1 &58.9 &49.8 &66.3 &69.5 &23.0 &84.6 &46.3 &55.8 \\
\metricname-L$_{\text{sup}}$ &56.7 &68.8 &59.2 &71.8 &70.0 &18.1 &83.1 &47.1 &58.0 &50.6 &65.3 &50.0 &66.6 &69.4 &12.3 &82.3 &45.7 &53.8 \\
\metricname-XL$_{\text{sup}}$ &54.2 &66.7 &63.0 &77.0 &73.2 &32.9 &87.7 &50.2 &62.6 &53.8 &67.8 &54.1 &74.4 &70.7 &34.4 &88.1 &55.7 &61.6 \\
\bottomrule
\end{tabular}
\end{table*}

\subsection{Dataset}
\subsubsection{Evaluation Dataset}
\label{sec:appendix-dataset-eval}
\paragraph{DA20} DA20 is the Direct Assessment (DA) corpus from WMT20 metrics shared task \citep{mathur2020results} which includes 23,293 segments across 18 language pairs. DA20 covers 18 language pairs and 211 systems:km-en(7), en-cs(12), pl-en(14), ru-en(11), iu-en(11), en-iu(11), ta-en(14), en-pl(14), en-ta(15), zh-en(16), en-zh(12), cs-en(12), de-en(12), ja-en(10), en-ja(11), en-de(14), ps-en(6), en-ru(9).

\paragraph{MQM20 \& MQM21} 
MQM20 \citep{freitag2021experts} and MQM21 \citep{freitag2021results} are datasets obtained by professional translators who re-annotated the outputs from WMT20 and WMT21 shared task according to the Multidimensional Quality Metrics (MQM) framework \citep{article}. The MQM framework contains assessments of five aspects of the text, which are accuracy, fluency, terminology, style, and local. MQM20 covers 2 language pairs and 20 systems: en-de(10), zh-en(10) and comprises 1,418 and 2,000 segments for language pair en-de, zh-en respectively. In our experiments, we excluded 3 human translation systems for en-de and 2 human translation systems for zh-en. MQM21 covers 2 language pairs and 32 systems: en-de(17), zh-en(15) including 527 and 650 samples for en-de, zh-en respectively. 


\paragraph{QE20}
QE20 \citep{specia2021findings} is the dataset of WMT20 shared task on Quality Estimation (QE). It covers 6 language pairs: en-de, en-zh, ro-en, et-en, ne-en, si-en, comprising 7,000/1,000/1,000 segments for Train/Dev/Test set for each language pair respectively, and each language pair uses one state-of-the-art machine translation models built using the fairseq toolkit (\url{https://github.com/facebookresearch/fairseq}).

\paragraph{MultiSumm}
MultiSumm \citep{koto2021evaluating} is a multilingual summarization dataset containing texts and their summaries in eight languages (en, id, fr, tr, zh, ru, de, es). The dataset collects 135 documents in each language, as well as summaries generated by 2 systems: Pointer-Generator \citep{see2017get} and BERT \citep{liu2019text, dong2019unified} model. 

\subsubsection{Parallel Dataset}
\label{sec:appendix-dataset-parallel}

\paragraph{ParaCotta} \cite{aji2022paracotta} A synthetic parallel paraphrase corpus across 17 languages. We use lexical BLEU between paraphrases to filter the dataset and keep the paraphrases with low lexical BLEU, in another word, high lexically diverse.\footnote{We keep the paraphrases with lexical BLEU in the range [0,20], while the possible range is [0,100].}  
\paragraph{MT-prism} \cite{thompson-post-2020-automatic} A Machine Translation dataset includes 99.8M training sentences across 39 languages. The data sources are WikiMatrix \cite{schwenk2019wikimatrix}, Global Voices,\footnote{https://casmacat.eu/corpus/global-voices.html} EuroParl \cite{koehn2005europarl}, SETimes\footnote{http://nlp.ffzg.hr/resources/corpora/setimes/}, United Nations \cite{eisele2010multiun}. 

\end{document}